\documentclass[10pt,journal,compsoc]{IEEEtran}
\usepackage{amssymb,amsmath}
\usepackage{graphicx}
\usepackage[ruled,linesnumbered]{algorithm2e}
\usepackage{booktabs}
\usepackage{multirow}
\usepackage{multicol}
\usepackage{color}
\usepackage{makecell}

\makeatletter
\renewcommand{\maketag@@@}[1]{\hbox{\m@th\normalsize\normalfont#1}}%
\makeatother

\definecolor{applegreen}{rgb}{0.55, 0.71, 0.0}

\definecolor{hycolor}{rgb}{0.7,0.7,0.3}

\ifCLASSOPTIONcompsoc
 \usepackage[nocompress]{cite}
\else
 \usepackage{cite}
\fi
\ifCLASSINFOpdf
\else
\fi
\begin{document}
%
\title{Collaborative Knowledge Graph Fusion by Exploiting the Open Corpus}

%
%
%
%

\author{Yue~Wang${}^{1}$,~Yao~Wan${}^{2}$,~Lu~Bai${}^{3}$,~Lixin~Cui${}^{1}$,~Zhuo~Xu${}^{1}$,~Ming Li${}^{4}$\\~Philip~S.~Yu${}^{5}$,~\IEEEmembership{Fellow,~IEEE}~and~Edwin~R~Hancock${}^{6}$,~\IEEEmembership{Fellow,~IEEE}
\thanks{Yue Wang, Lu Bai (${}^{*}$Corresponding Author: bailucs@cufe.edu.cn), Lixin Cui, and Zhuo Xu are with ${}^{1}$Central University of Finance and Economics, Beijing, China. Yao Wan is with ${}^{2}$College of Computer Science and Technology at Huazhong University of Science and Technology (HUST), Wuhan, China. Lu Bai is with ${}^{3}$School of Artificial Intelligence, Beijing Normal University, Beijing, China. Ming Li is with ${}^{4}$the Key Laboratory of Intelligent Education Technology and Application of Zhejiang Province, Zhejiang Normal University, Jinhua, China. Philip S. Yu is with ${}^{5}$Department of Computer Science, University of Illinois at Chicago, US. Edwin R. Hancock is with ${}^{6}$Department of Computer Science, University of York, UK. This work is supported by the National Natural Science Foundation of China under Grants T2122020, 61976235, and 61602535. This work is also supported in part by NSF under grants III-1526499, III-1763325, III-1909323, and CNS-1930941.}
}

%
%

\markboth{IEEE Transactions on Knowledge and Data Engineering,~Vol.~14, No.~8, August~2020}%
{Yue Wang \MakeLowercase{\textit{et al.}}: IEEE Transactions on Knowledge and Data Engineering}
%




\IEEEtitleabstractindextext{%
\begin{abstract}
To alleviate the challenges of building Knowledge Graphs (KG) from scratch, a more general task is to enrich a KG using triples from an open corpus, where the obtained triples contain noisy entities and relations. It is challenging to enrich a KG with newly harvested triples while maintaining the quality of the knowledge representation. This paper proposes a system to refine a KG using information harvested from an additional corpus. To this end, we formulate our task as two coupled sub-tasks, namely join event extraction (JEE) and knowledge graph fusion (KGF). We then propose a Collaborative Knowledge Graph Fusion Framework to allow our sub-tasks to mutually assist one another in an alternating manner. More concretely, the explorer carries out the JEE supervised by both the ground-truth annotation and an existing KG provided by the supervisor. The supervisor then evaluates the triples extracted by the explorer and enriches the KG with those that are highly ranked. To implement this evaluation, we further propose a Translated Relation Alignment Scoring Mechanism to align and translate the extracted triples to the prior KG. Experiments verify that this collaboration can both improve the performance of the JEE and the KGF.
\end{abstract}

}

\newtheorem{example}{Example}[section]
\maketitle

\IEEEdisplaynontitleabstractindextext

%
\IEEEpeerreviewmaketitle

\IEEEraisesectionheading{\section{Introduction}\label{sec:introduction}}
\IEEEPARstart{K}{nowledge} graphs, which are a structurally organized form of information,
have supported a variety of downstream tasks,
including recommender systems~\cite{DBLP:conf/kdd/Wang00LC19}, NLP tasks~\cite{annervaz2018learning}, question answering~\cite{DBLP:conf/naacl/TalmorB18,8085196}, and entity-linking~\cite{9384305}.
Existing open source knowledge graphs such as Wikidata~\cite{DBLP:journals/cacm/VrandecicK14}, WordNet~\cite{DBLP:journals/cacm/Miller95} and Freebase~\cite{DBLP:conf/sigmod/BollackerEPST08} contain billions of Resource Description Framework (RDF) triples~\cite{DBLP:journals/semweb/FarberBMR18} in the form of \textit{(subject, relation, object)} relations, where both the \textit{subject} and \textit{object} represent the named entities~\cite{9039685}, and the \textit{relation} models the relationship between these two named entities. However, since open source knowledge graphs 
are designed for general purposes, they contain only limited factuaL knowledge 
for particular tasks~\cite{DBLP:conf/cikm/LiuBLZSWX19} in restricted domais such as finance or medicine. To adapt to multiple domains, it is crucial to construct high-quality domain specific knowledge graphs.

In order to construct new knowledge graphs from unstructured textual sources, existing work mainly consists of several pipelined sub-tasks, e.g., named entity recognition~\cite{DBLP:conf/naacl/LampleBSKD16}, relation extraction~\cite{DBLP:conf/acl/LinSLLS16} or relation alignment~\cite{DBLP:conf/esws/KoutrakiPV17}.
These methods are designed as separate subtasks and not as a unified system~\cite{DBLP:journals/www/ZhaoJLJS20}. Thus they do not fully address the issue of how to effectively leverage the information hidden in the connections between the subtasks~\cite{wang2020crosssupervised} to improve the quality of a knowledge graph built from a text corpus. 
To this end, recent work has combined named entity recognition with relation extraction as a single joint-event-extraction~\cite{huang-etal-2020-joint-event} task that can jointly obtain the entities and relations from text sources. However, since the current work does not focus on the resulting process to build an integrated knowledge graph from the extracted results, there still exists much scope for constructing a high-quality domain-oriented knowledge graph from test documents.

Knowledge graph fusion~\cite{DBLP:journals/inffus/NguyenVJ20,DBLP:conf/kdd/0001GHHLMSSZ14,DBLP:journals/www/ZhaoJLJS20} is a possible route by which to construct a knowledge graph from the extracted event factors in an open corpus. Early work applied the traditional data fusion method~\cite{10.1145/2723372.2731083} while considering only fusing the data under a global or compatible data schema~\cite{10.1145/1456650.1456651}. This work evaluates the quality of data by checking whether or not a triple is contained in the extended set of a ground-truth knowledge graph~\cite{10.14778/2732951.2732962}. However, this type of method may ignore the
implications of knowledge that is indirectly contained in the ground-truth knowledge graph. It may thus discard many meaningful triples from different and potentially valuable sources. In order to overcome this problem, recent knowledge graph embedding~\cite{sourty-etal-2020-knowledge} methods have leveraged network embedding technology~\cite{DBLP:journals/tkde/CuiWPZ19} to infer the possibilities of the existence of triples in a given knowledge graph. This is done by representing the triples as latent vectors~\cite{DBLP:conf/nips/SocherCMN13,DBLP:conf/ijcai/WangWG15,DBLP:conf/cikm/GuanJWC18}. Specifically, with the representation vectors of the triples to hand, these methods use statistical models~\cite{DBLP:conf/nips/BordesUGWY13} or neural networks~\cite{dettmers2018conve,Nguyen2018} to predict plausible scores for the potential triples.

\begin{figure*}[ht]
	\centering
		\includegraphics[width=6.1in]{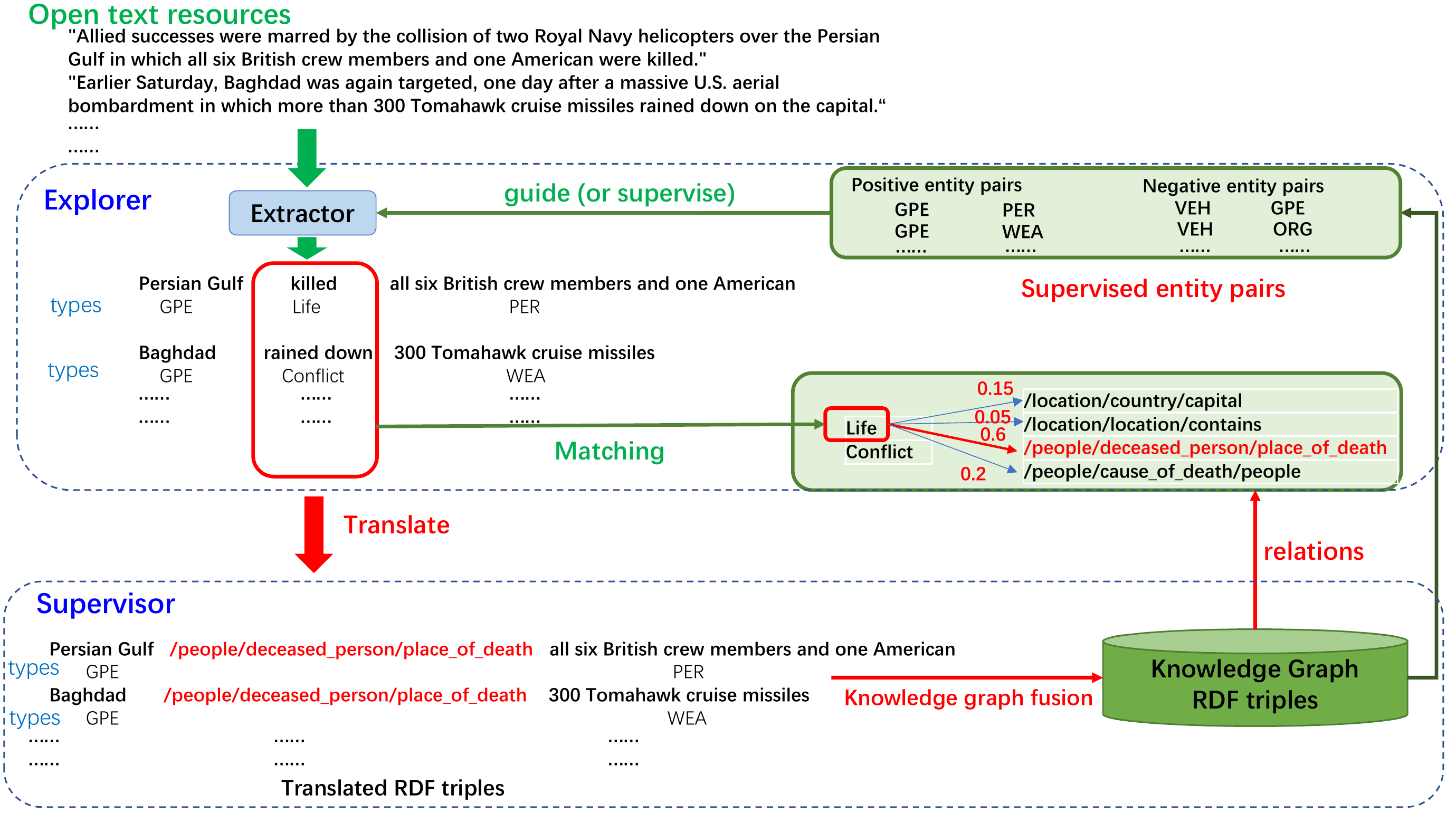}
	\caption{In a collaborative knowledge graph fusion process, an explorer and a supervisor collaborate to create an enriched knowledge graph by extending a prior knowledge graph with RDF triples extracted from open text sources. Since the extracted RDF triples contain entities or relations that are not aligned to the prior knowledge graph, this process requires interaction mechanisms (translate the extracted results to the knowledge graph RDF triples and guide the explorer with meaningful entity pairs) between the explorer and the supervisor. To simplify the problem, we suppose both the explorer and supervisor share the same entity types (Geographical/Social/Political Entities (GPE), Persons (PER), Weapons (WEA), Organizations (ORG), Vehicles (VEH), etc.) and the extracted trigger mentions (killed, rained down, etc.) by the explorer belong to the trigger types (Life, Conflict, etc.) by following the definitions in the ACE 2005 corpus~\cite{ldc}. Then the core problem becomes to align the trigger mentions obtained by the explorer to the relations in the knowledge graph of the supervisor.}
	\label{fig:motivation} 
  \vspace{-0.1in}
\end{figure*}

Although much existing work discusses the potential triple evaluation problem for the knowledge graph fusion task, little considers generating the candidate triples from open text sources and linking candidate generation with the evaluation process to automatically. fuse the obtained triples to a prior knowledge graph. The main challenges that hinder progress in this direction are routed in the following shortcomings in the knowledge extraction and a knowledge graph fusion tasks.
(1) \textit{Difficulties in aligning RDF triples}. Since open text sources may contain relations outside the scope of a prior knowledge graph, it is a challenge to align the relations from the open texts to those in the knowledge graph.
Although current work discusses the entity alignment~\cite{DBLP:conf/aaai/TrisedyaQZ19} between sources, little focusses on relation alignment.
This leads to the difficulty of aligning the extracted RDF triples from the text sources to a prior knowledge graph.
(2) \textit{Difficulties maintaining knowledge graph quality}.
Merging the unaligned RDF triples from the open text sources to a knowledge graph can mislead the knowledge graph embedding model and may result in unreliable plausible scores for potential triples. Moreover, a misleading knowledge graph can result in the the extractor 
relying on low-quality triples. This may further lower the quality of the knowledge graph. (3) \textit{Difficulties sharing knowledge between sub-tasks.} Without a reliable way of aligning the RDF triples, it becomes difficult to share knowledge between the sub-tasks (e.g. event extraction and knowledge fusion). This leads to error propagation~\cite{DBLP:conf/emnlp/ZengLC015} between sub-tasks and thus degrade the performance for each sub-task.

To address the aforementioned limitations, in this paper, we formulate a new method that combines event extraction (extractor) with knowledge graph fusion as a Collaborative Knowledge Graph Fusion process. 
Specifically, we propose a unified framework to build a domain-oriented knowledge graph by enriching an open-source knowledge graph with knowledge extracted automatically from a text corpus. 
Since our new method provides a mechanism to share the knowledge between sub-tasks, our enriched knowledge graph grows larger by incorporating facts of knowledge from the texts. In addition, the new method also leverages the enriched knowledge graph to assist our event extraction sub-task to obtain more reliable entities and relations from documents.

As illustrated in Figure~\ref{fig:motivation}, the collaborative knowledge graph fusion method consists of two interacting processes, an explorer and a supervisor. That is, by referring to the principles (e.g. the possible entity pairs) from a supervisor, an extractor explores new RDF triples from the available open text sources. After the extractor submits the newly discovered triples to the supervisor, the supervisor evaluates their quality and extends the existing set of triples using the highest quality newly discovered triples.

Specifically, our framework guides the extractor with the entity pairs from a prior seed knowledge graph, and then iteratively increments the seed knowledge graph with the extracted triples from the extractor. 
In this process, both the performance of the extractor and the quality of the enriched knowledge graph are improved. 
To this end, in our extractor, we propose a benchmark-based supervision mechanism to supervise the extraction process with the entity pairs from the seed knowledge graph maintained by the supervisor. 
This is implemented by a contrastive learning method which considers both the positive and negative entity pairs.
These entity pairs are sampled from the prior knowledge graph with a neural Knowledge Graph Embedding (KGE) scoring function trained by the supervisor process. 
On the other hand, to the supervisor, the KGE scoring function is trained by the triples in the seed or the enriched knowledge graph and it evaluates the matching degree of the extracted RDF triples from the extractor to the knowledge of the supervisor. 
Consequently, the supervisor merges the high-ranked triples from the extracted results into the prior knowledge graph.

We conduct exhaustive experiments on real-world corpora and knowledge graphs. Experimental results show that our system achieves higher performance than state-of-the-art baselines, both on the joint-event-extraction and the knowledge-graph-embedding tasks. 
This verifies not only that the proposed benchmark-based supervision mechanism guides the extractor well in our system, and but that it also implies that the knowledge graph of the supervisor maintains a high quality by being enriched with the triples evaluated by the supervisor.

In summary, our main contributions are as follows:
\begin{itemize}
  \item We formalize the knowledge graph fusion with open corpora as an alternating process consisting of extracting the RDF triples from documents and then fusing a prior knowledge graph with the obtained triples. As far as we know, our work is the first to discuss a unified architecture to conduct the knowledge fusion directly based on the text sources. 
  \item We propose the ``Collaborative Knowledge Graph Fusion'' framework as a solution for the aforementioned problem. In this framework, we propose the Benchmark-based Supervision Mechanism to further supervise the performance of our JEE process (in the explorer process) with positive and negative entity pairs sampled from a prior KG provided by the supervisor. 
  \item We propose an unsupervised metric, Translated Relation Alignment Scoring (TRAS), to assist align and translate the extracted triples from the JEE process to those in the proper form to the prior KG.
  \item With the proposed Benchmark-based Supervision Mechanism and TRAS to hand, we implement the ``Collaborative Knowledge Graph Fusion'' as a unified process. It automatically extracts the triples from an open corpus and enriches them to a given prior KG in an alternative process.
  \item Our experiments on several real-world datasets show that, with the proposed framework, our system achieves better performance both on the JEE and KGF tasks than the related alternatives. This verifies that our method not only improves the JEE process but also yields a high-quality enriched KG. Specifically, our case study shows that our system could translate the extracted triples from a text corpus to the facts consistent with a prior KG with the assistance of the proposed TRAS score. This improves the quality of the prior KG and also explains the reason for the performance improvement of the KGF task.
\end{itemize}

The remainder of this paper is organized as follows. In Section~\ref{sec:preliminaries}, we introduce the preliminaries concerning the joint event extraction and knowledge graph fusion processes and then also formalize the problem of knowledge graph fusion with an open corpus. Section~\ref{sec:framework} presents in detail our proposed framework and fusion mechanism. Section~\ref{sec:experiment} verifies the effectiveness of our model and compares it with recent methods on real-world datasets. Section~\ref{sec:related_works} summarizes recent related work. Finally, we conclude this paper in Section~\ref{sec_conclusion} where we offer suggestions for further work in this direction.

\section{Preliminaries}\label{sec:preliminaries}

Our overall objective is knowledge graph fusion with an open corpus. This task consists of a joint event extraction (JEE) step to extract knowledge triples from unstructured texts and a knowledge graph fusion (KGF) step to evaluate and enrich the extracted triples from the JEE step for a prior or exsiting Knowledge Graph (KG). 
We elaborate the notation for the JEE and KG, and formalize our problem in the following subsections.

\subsection{Knowledge Graphs}

A Knowledge Graph (KG)~\cite{DBLP:journals/corr/abs-2002-00388} is represented as a set of factual (RDF) triples referring to specific topics. Formally, we define a knowledge graph $G$ in the structure $G = \langle{E,R,T}\rangle$, where $E$ is a set of entities, $R$ is a set of relations and $T$ is the set of the RDF triples. For example, $G_1 = \langle{E_1,R_1,T_1}\rangle$ is a knowledge graph
of  capital city relationships with the entity set $E_1=\{Tokyo, Beijing, Japan, China\}$, the relation set $R_1=\{capital\_of\}$ and the triple set $T_1=\{\langle{Tokyo, capital\_of, Japan}\rangle,\langle{Beijing, capital\_of, China}\rangle\}$. Since a human-composed document does not contain such structural information such as the entities, relationships or triples, to build a KG from a corpus, we require to extract the triples from the texts. 

\subsection{Joint Event Extraction}
\label{sec:JEE}
Event extraction is a technique to extract the structural information such as entities or relations~\cite{DBLP:conf/naacl/LampleBSKD16} from a given corpus. This requires applying sub-tasks such as Named Entity Recognition (NER) and Relation Extraction (RE). 
Traditional methods train separate multi-label classifiers to distinguish the labels for the tokens (both for the entity and text relation mentions) in sentences.
In order to improve the accuracy of the extraction process, recent work leverages the pipelined method to classify the relationship first and then identify the entities with roles centered around the determined relation. However, since these methods invoke their sub-processes separately, they
 feedback weakly from the entity identification task to the preceding tasks. As a result they may
 suffer from limitations caused by error-propagation~\cite{DBLP:conf/naacl/YangM16}.

To this end, we use a universal sequence-to-sequence (Seq2Seq) framework~\cite{wang2020crosssupervised} to simultaneously extract the entities and relations from a text corpus. 




\textbf{Seq2Seq Joint-Event-Extraction (JEE).}
Let the text corpus $D$ be a set of sentences, where $D=\{s_1, s_2, s_3, \ldots\}$ ($\forall{s}\in{D}$, $s=\{w_1,w_2,w_3,\ldots,w_m\}$, where $w_i$s are tokens). Let $\mathcal{A}=\mathcal{A}_E\bigcup\mathcal{A}_R$ be a combined tag set with predefined types for tokens, where $\mathcal{A}_E$ and $\mathcal{A}_R$ are the sets of the predefined entity and text relation mention types respectively. Then the aim of JEE is to find an optimal map $\mathcal{Y}_{\Theta{_1}}:s\rightarrow{\Pi_{i=0}^{M}\mathcal{A}}$, ($\forall{s}\in{D}$), where $\Pi$ is the Cartesian product, $M$ is the maximum length for the sentences in $D$, $\Theta{_1}$ is the vector for the learned parameters.

In this form, our JEE process transforms a sentence into a tag sequence with the tags in the combined tag set $\mathcal{A}$. The loss function for the Seq2Seq JEE is computed as a cross-entropy function, as follows:
\begin{equation}\label{eq:jee}
  \mathcal{L}_{jee}=\sum_{i=0}^{M}\sum_{y_i\in\mathcal{A}}-Pr(y_i|w_i)\log{\hat{Pr}(y_i|w_i)}.
\end{equation}
With the mapped tag sequence optimized by the loss function in Equation~\ref{eq:jee}, we obtain the annotated tag sequences for the sentences in a corpus. In this manner, the entity and relation mentions for a sentence are extracted together. Consequently, we generate RDF triples based on their extracted mentions and use these triples as the candidate triples for KG enrichment. In order to simplify the discussion, we use the term $\mathcal{Y}_{\Theta{_1}}$ as a joint operation that combines both the mapping from sentences to label sequences and the RDF generation process. Therefore, $\mathcal{Y}_{\Theta{_1}}(D)$ refers to a set of RDF triples and we refer to it as the {\it extractor map} in the following sections.

\subsection{Knowledge Graph Fusion with an Open Corpus}\label{sec:KGF_definition}

Knowledge Graph Fusion~\cite{DBLP:journals/inffus/NguyenVJ20} is the task of constructing a unified knowledge graph from different data sources. Traditional knowledge graph fusion aims to integrate several knowledge graphs into one knowledge graph, and we formalize this task as follows:

\textbf{Knowledge Graph Fusion (KGF).} Given two prior knowledge graphs $G_1=\langle{E_1,R_1,T_1}\rangle$ and $G_2=\langle{E_2,R_2,T_2}\rangle$, suppose both $G_1$ and $G_2$ are used under the same RDF schema to build a new knowledge graph $G'=\langle{E',R',T'}\rangle$, where $T'=T_1\bigcup{\Delta{T}}$ and $\Delta{T}$ is the set of triples of $G_2$ with the top-K plausible scores $ f_{G_1}(i,r,t)$ ($\forall{(i,r,t)}\in{G_2}$). This score is computed as
\begin{equation}
\label{eq:plausible}
  f_{G_1}(i,r,t)=\sum_{(i^*,r^*,t^*)\in{T_1}}Sim((i,r,t),(i^*,r^*,t^*)),
\end{equation}
where the function $Sim$ gives the similarity between two triples. The plausibility score of a triple evaluates the consistency of this triple with an existing or prior knowledge graph. Since it is inefficient to compute the plausibility score by traversing all the triples of a knowledge graph, mainstream work applies the Knowledge Graph Embedding (KGE)~\cite{sourty-etal-2020-knowledge} method for this evaluation. Specifically, these methods generate the vector representations for triples and compute the similarities between triples through their vector similarities. Recent methods represent the knowledge triple as latent vectors by following the ideas introduced in the translation based embedding model (TransE)~\cite{DBLP:conf/nips/BordesUGWY13}.


\textbf{Knowledge Graph Embedding (KGE).} Given a KB $G=\langle{E, R, T}\rangle$, suppose $(i,r,j)$ is a triple from $T$, then the loss is 
\begin{equation}
\label{eq:kge:loss}
\mathcal{L}_{kge}\!=\!-\!\sum_{(i,r,j)\in{T},\atop(i',r,j')\in{N}}\!||\gamma+f_G(i, r,j))-f_G(i',r,j')||
\end{equation}
where $N$ is the corresponding negative set for the triples in $T$, $\gamma$ is a hyperparameter, $f_G(i,r,j)$ is a scoring function to evaluate the consistency of any triple ($i,r,j$) to the knowledge graph G and the normalization in Equation~\ref{eq:kge:loss} can be based on either the L1 or L2-norm. According to the design of TransE, a plausibility score $f_G(i,r,j)$ can be computed as the following.
\begin{equation}
\label{eq:kge:scoring_transe}
f_G(i,r,j)=d(e_i+e_r,e_j),
\end{equation}
where $e$ is an embedding that maps any entity or relation to an $\mathbb{R}^h$ vector and $d(*,*)$ is the Euclidean distance function between two $\mathbb{R}^h$ vectors.

Therefore, with a trained embedding $e$ based on the given prior knowledge graph $G_1$, the plausibility of a triple $(i,r,j)$ from $G_2$ to $G_1$ can be evaluated by computing the Euclidean distance $d(e_i+e_r,e_j)$.

As discussed in the Introduction, our objective is to build a knowledge graph fusion system using open text sources. This task is different from the aforementioned knowledge graph fusion and it means we require to: (1) extract the RDF triples from a given corpus $D$ and (2) fuse the extracted triples to a knowledge graph $G$. Specifically, we formalize this problem as the following.


\textbf{Open Knowledge Graph Fusion (OKGF).} 
Given a prior knowledge graph $G=\langle{E, R, T}\rangle$, a corpus $D$ and an extractor map $\mathcal{Y}_{\Theta{_1}}$, suppose $\mathcal{Y}_{\Theta{_1}}(D)$ is a set of extracted triples from a corpus $D$. Then with a trainable scoring function $f(*)$ and embedding map $e$, the objective of OKGF is to find the optimal subset $\Delta{T}$ from $\mathcal{Y}_{\Theta{_1}}(D)$ that minimizes the following loss function:
\begin{equation}
\label{eq:problem1}
  \mathcal{L}_{OKGF}\!=\!-\!\sum_{(i,r,j)\in{T\bigcup{\Delta{T}}},\atop(i',r,j')\in{N}}\!||\gamma+f_G(i, r,j))-f_G(i',r,j')||,
\end{equation}
where $N$ is the corresponding negative triple set for the positive triples $t$ from $T$.

This task links the JEE and the KGF processes together. However, it is a combinatorial optimization problem that exhaustively checks all the possible subsets $\Delta{T}$ from $\mathcal{Y}_{\Theta{_1}}(D)$. The newly discovered noisy entities and relations from the open corpus exacerbate the problem. Therefore, it is difficult to obtain the global optimal solution. To this end, we propose a heuristic collaborative knowledge graph fusion framework to connect the JEE and KGF subtasks to fuse an open corpus to obtain a prior knowledge graph. Our framework approaches open knowledge graph fusion 
from two directions, namely 1) our model guides the JEE process with a prior knowledge graph and 2) it selectively enriches the prior knowledge graph with the extracted results from the JEE process. This requires a careful design of both the JEE supervision mechanism with a prior knowledge graph and an effective ``translation-and-evaluation'' method to fuse the extracted results into the prior knowledge graph. We elaborate the details in the next section.

\section{Our Proposed Method}\label{sec:framework}

In this section, we introduce the Collaborative Knowledge Graph Fusion framework to address knowledge graph fusion with an open corpus. 
\subsection{Overview}
\begin{figure*}[htb]
	\centering
		\includegraphics[width=7.1in]{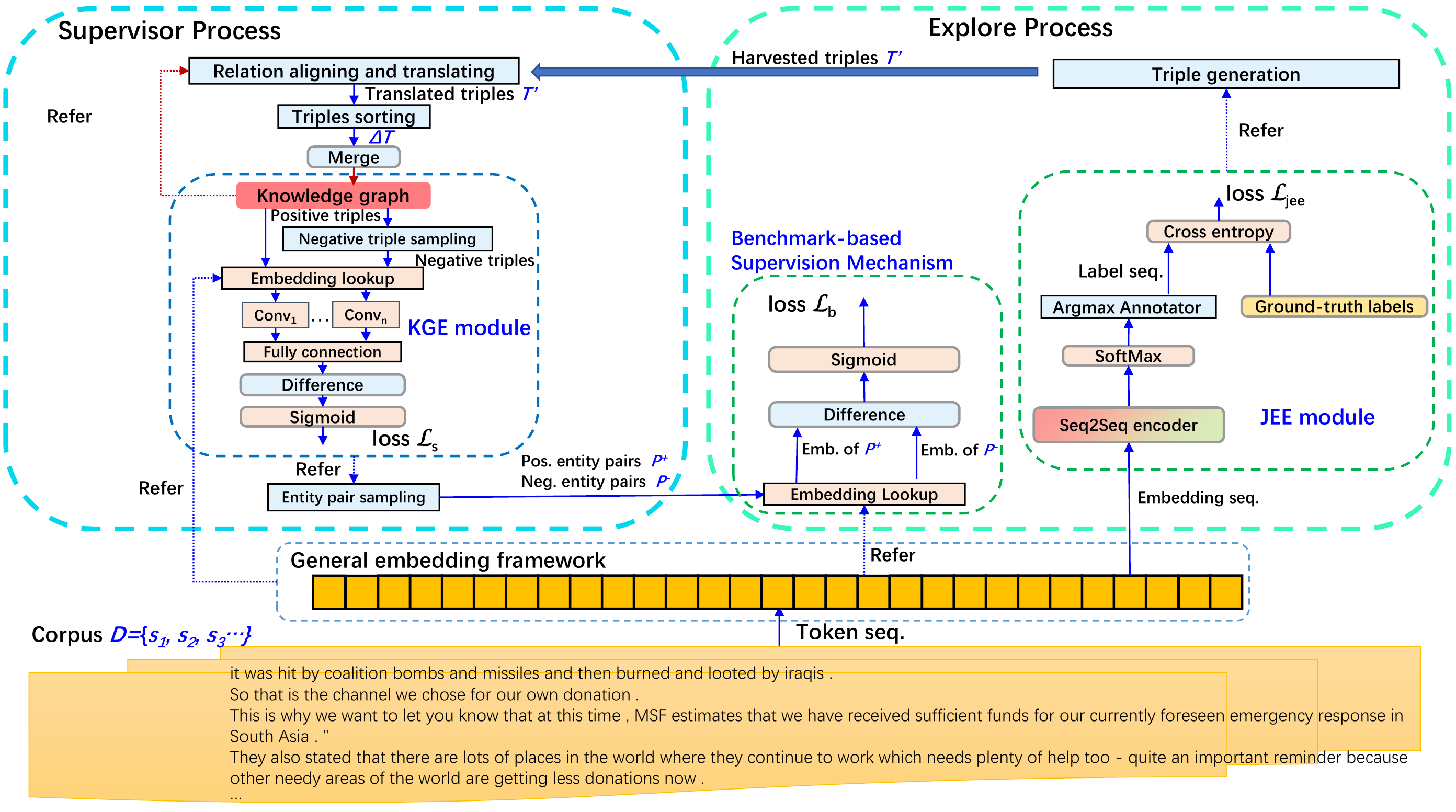}
	\caption{The ``Collaborative Knowledge Graph Fusion'' framework for the Knowledge Graph Fusion with Open Corpus task. Our framework consists of two alternative running processes: 1) an explorer process carries on the Joint-Event-Extraction (JEE) task and 2) a supervisor process aligns and merges the extracted triples to a prior knowledge graph. Our system first embeds the texts to the latent vectors of tokens and then optimizes the forward scores for the explorer process. After training the JEE model, our system extracts the triples $T'$ from the open texts. Then, our system treats them as candidate triples and enriches them to the prior KG by referring the proposed Translate Relation Alignment Score (TRAS). The enriched KG and the trained KGE likelihood scoring function helps to sample the top positive and negative entity pairs for the explorer process in return.}
	\label{fig:framework} 
  \vspace{-0.1in}
\end{figure*}

To emulate a human-like collaborative process for our task, we propose a system with two processes, namely 1) an explorer process and 2) a supervisor process. In the explorer process, the system uses the proposed Benchmark-based Supervision Mechanism to assist the JEE task to extract the triples while guided by a supervisor (the benchmarks discovered by the supervisor from a prior KG). In the supervisor process, the system applies the proposed Relation Alignment-based Knowledge Graph Fusion module to selectively accept the extracted triples to be added to the prior KG. These two processes alternate to simultaneously extract knowledge triples and enrich a prior KG with high-quality. Figure \ref{fig:framework} illustrates the architecture of our system. The details for the proposed processes are given in the following subsections.

\subsection{The Explorer: Benchmark-based Supervision JEE}
\label{sec:explorer}

In Figure~\ref{fig:framework}, our explorer process implements the JEE task. To ensure the
explorer is guided by the supervisor we introduce a Benchmark-based Supervision Layer. In this work, we apply the Seq2Seq JEE as the basic extraction process and use BERT~\cite{DBLP:conf/naacl/DevlinCLT19} as the sequence-to-sequence encoder. This JEE module can be substituted by any alternative JEE model if necessary.

Intuitively, during the exploratory period, an explorer receives examples from a supervisor and attempts to leverage the knowledge in these examples to facilitate better exploration. In our work, the explorer process extracts the triples from an open corpus based on a prior KG maintained by a supervisor. Since the open corpus may contain unaligned relations and extra entities that are not contained in the prior KG, it requires a relatively flexible method rather than strict supervision to guide the explorer. To this end, we introduce 
the Benchmark-based Supervision Mechanism.

\textbf{Benchmark-based Supervision Mechanism.} Given a prior KG, $G=\langle{E,R,T}\rangle$, let the benchmarks be a positive set of entity pairs $P^+$ and a negative set of entity pairs $P^-$, where $P^+=\{(i,j)|(i,*,j)\in{T}, \forall{i,j}\in{E}\}$, and $P^-=\{(i,j)|(i,*,j)\notin{T}, \forall{i,j}\in{E}\}$. Then the Benchmark-based Supervision Mechanism can be described as the task to minimize a loss function extended from the BPR loss~\cite{https://doi.org/10.48550/arxiv.1205.2618}
\begin{equation}
\label{eq:kg_guide}
  \mathcal{L}_b=-\log{(\delta(f(P^+)-f(P^-))},
\end{equation}
where $\delta$ is the Sigmoid function, $f(P)$ is a function to compute the likelihood for any entity pair ($i$,$j$) ($\forall{(i,j)}\in{P}$), and is given by 
\begin{equation}
\label{eq:fs}
  f(P)=\rm{ffnn}(\sum_{\forall{i,j}\in{P}}(e_i-e_j)),
\end{equation}
where $e_i$ is an $\mathbb{R}^d$ embedding vector for any entity $i$ ($\forall{i}\in{E}$); ``ffnn'' is a fully connected neural network to map an $\mathbb{R}^d$ embedding vector to an $\mathbb{R}^1$ score.

Optimizing $\mathcal{L}_b$ results in the training of a scoring function $f(P)$ to measure the likelihood of any entity pair while maximizing the difference between the likelihood scores of the positive and negative entity pairs. This fits with the intuition that an explorer understands the knowledge in the examples from the supervisor.

Further, since an entity is a sequence of tokens with arbitrary lengths, we apply the weighted average method \cite{DBLP:conf/iclr/AroraLM17} to represent an entity by its corresponding embedding vector. Formally, the embedding vector for an entity is computed as follows
\begin{equation}
\label{eq:entity_emb}
  e_i = \sum_{\forall{w}\in{i}}e_w,
\end{equation}
where $i$ is an
entity in $E$ and $w$ is any token in the entity $i$. The embedding vector $e_w$ can be obtained by referring to the embedding dictionary table.

With the proposed Benchmark-based Supervision Mechanism, the loss function of our explorer process is a weighted sum of Equations \ref{eq:jee} and \ref{eq:kg_guide}, i.e. 
\begin{equation}
\label{eq:gjee}
  \mathcal{L}_e = (1-\alpha)\mathcal{L}_{jee}+\alpha\mathcal{L}_b,
\end{equation}
where $\alpha$ is the weight for the benchmark-based supervision.

\textbf{Candidate Triple Set.} With the aforementioned explorer process, our system simultaneously extracts the entity and relation mentions (or triggers). Then, we generate all RDF triples exhaustively based on the extracted mentions. The results are treated as the candidate triple set $T'$ for subsequent processing steps.

\subsection{The Supervisor: Relation Alignment-based OKGF}
\label{sec:supervisor}

Our supervisor process enriches the prior KG with the optimal subset of the candidate triples from the explorer process. This requires a scoring function to measure the plausibilities for triples trained by the prior KG. The process for a supervisor to evaluate the quality of the discovery is similar to that adopted 
by the explorer. As is discussed in Section \ref{sec:KGF_definition}, one of the challenges to implementing this task is that the relation mentions from the candidate triples may not be unaligned to the relations in the prior KG. In order to address this issue, we propose the Translated Relation Alignment Score (TRAS). This score facilitates the alignment of the relations between the candidate triples and the existing relations in the prior KG. After aligning the relations, our system translates the candidate triples to the aligned candidate triples. It then ranks the aligned candidate triples by considering the semantic information residing in the prior KG. The highly-ranked triples are integrated into the prior KG to generate an enriched KG. We expand the details of this process in the remainder of this section.


\textbf{Translated Relation Alignment Score (TRAS).} Given two KGs $G_1=\langle{E_1,R_1,T_1}\rangle$ and $G_2=\langle{E_2,R_2,T_2}\rangle$ sets ($T_1\bigcap{T2}=\phi$). Then the TRAS score $s(r_1,r_2)$ between two relation $r_1$ and $r_2$ ($\forall{r_1}\in{R_1}, \forall{r_2}\in{R_2}$) is computed as follows
\begin{equation}
\label{eq:tras}
  s(r_1,r_2)=\gamma{s_m(r_1,r_2)} + (1-{\gamma})s_e(r_1,r_2),
\end{equation}
where $s_m(r_1,r_2)$ is the text mention similarity between $r_1$ and $r_2$, $\gamma$ is the weight of the text mention similarity. The quantity 
$s_e(r_1,r_2)$ is the \textbf{translated relation similarity} between two relations ($r_1$ and $r_2$) which can be computed
as follows
\begin{equation}
\label{eq:sim}
  s_e(r_1,r_2)=Sim(\sum_{\forall(i,r_1,j)\in{T_1}}{e_i-e_j},\sum_{\forall(i,r_2,j)\in{T_2}}{e_i-e_j}),
\end{equation}
where $Sim(*,*)$ can be any similarity function between two vectors. In this paper, we use the Cosine similarity for this task. Generally, the summed entity embedding difference in Equation \ref{eq:sim} represents the embedding vector for a given relation. As a result, Equation \ref{eq:sim} computes the proximity between two relations in different KGs by considering the entities adjacent to them. 

\textbf{Aligned Triple Set.} Our system ranks the relation pairs between the candidate triples from $T'$ and the triples in the prior KG using their TRAS scores. As a result, our system translates the candidate triples from the JEE process to an aligned triple set with the same relation set in the prior KG. The aligned triple set is denoted by $\Delta{T}$.

\textbf{Knowledge Graph Embedding (KGE) Triple Likelihood.} After generating the aligned candidate triple set from the extracted triples, the supervisor ranks the candidate triples and merges the top-ranked triples to the current prior KG. To this end, we use a Knowledge Graph Embedding (KGE) Triple likelihood to perform the ranking task for triples. This function represents the action of the supervisor and it is implemented using 
a Convolutional Neural Network (CNN)~\cite{DBLP:conf/nips/HuLLC14} based model to map the triples to an $\mathbb{R}^1$ score. Formally, given a KG $G=\langle{E,R,T}\rangle$. The KGE triple likelihood $f_G(i,r,j)$ ($\forall(i,j)\in{E},\forall{r}\in{R}$) is computed as follows
\begin{equation}
\label{eq:kgelikelihood}
  f_G(i,r,j)=\delta(F([C_1,C_2,C_3,\ldots, C_{m}])),
\end{equation}
where $\delta$ is the Sigmoid function, $F$ is a fully-connected layer to map the concatenated convolution results to a $\mathbb{R}^{1}$ score that refers to the plausible probability for the triple ($i,r,j$) based on $G$. The quantity $C_n$ is the $n$-th convolutional result which can be computed as follows
\begin{equation}
\label{eq:kge_c}
  C_n=Maxpool(Relu(W_n\otimes[e_i^T,e_r^T,e_j^T]+B_n)),
\end{equation}
where $W_n$ is the $n$-th ($n$=$1$, $2$, \ldots, $m$) convolutional kernel and $B_n$ is the corresponding bias, $\otimes$ is the convolution operator, $Maxpool$ is the Maxpooling function, $Relu$ is the ReLU active function and $e_r$ is the embedding vector for the relation $r$. To alleviate the problems of sparsity in the extracted relations, rather than the one-hot encoding with a fixed dictionary, we applied a similar method to Equation \ref{eq:entity_emb} to sum all the tokens in a relation mention to obtain the embedding vector $e_r$ of a relation $r$.

The KGE triple likelihood is trained by optimizing a BPR loss function

\begin{equation}
\label{eq:kge_loss}
  \mathcal{L}_s\!=\!-\sum\limits_{\forall{(i,r,j)}\in{T\bigcup{\Delta{T}}},\atop\forall{(i',r,j')}\in{N}}\!\log{(\delta\!\left(\!f_G(i,r,j)\!-\!f_G(i',r,j')\!\right)\!)}.
\end{equation}
Optimizing this loss function maximizes the difference between the positive and negative triples. Since this training uses all of the triples in the prior KG, the trained KGE triples likelihood represents the action of a supervisor based on the current KG.

\textbf{Benchmark Entity Pairs Sampling}.
With the KGE triple likelihood to hand, we propose an algorithm (cf. in Algorithm \ref{algorithm:sampling}) to obtain the top positive and negative set pairs based on the current KG and embedding.

\begin{algorithm}\small
\label{algorithm:sampling}
\DontPrintSemicolon \KwData{a KG $G=\langle{E,R,T}\rangle$, the embedding mapper $\mathcal{E}$ from the JEE process, a threshold $k$.}
\KwResult{the positive entity pair set $P^+$, the negative entity pair set $P^-$.}
\Begin {
Compute all $f_G(i,r,j)$s ($\forall{(i,r,j)\in{T}}$) with Eq. (\ref{eq:kgelikelihood}).\\
Sort the triples in $T$ in ascending order and select the top-k ranked entity pairs $P^+$.\\
Enumerate all the negative triples $N$ ($\forall{(i,r,j)\notin{T}, \forall{i,j}\in{E}, \forall{r}\in{R}}$).\\ 
Compute all $f_G(i,r,j)$s ($\forall{(i,r,j)\in{N}}$) with Eq. (\ref{eq:kgelikelihood}).\\
Sort the triples in $T'$ in descending order and select the top-k ranked entity pairs $P^+$.\\
\textbf{Output} $P^+$ and $P^-$.\\
}
\caption{Benchmark Entity Pairs Sampling}
\end{algorithm}
The sampled positive and negative entity pairs are used directly as the benchmarks to supervise the explorer process (cf. Equation \ref{eq:kg_guide}). This simulates the way in which the supervisor provides the key examples to the explorer for the exploration task.

\subsection{The Complete Process and Discussion}
The complete Collaborative Knowledge Graph Fusion process is described in the Algorithm \ref{algorithm:em}. 
\begin{algorithm}\small
\label{algorithm:em}
\DontPrintSemicolon \KwData{A prior KG $G=\langle{E,R,T}\rangle$,
a corpus $D$ and a threshold $k$ for the polarity triple sampling and a threshold $\varepsilon$ for the KG enrichment.}
\KwResult{An enriched KG $G'$.}
\Begin {
Initialize the embedding mapper $\mathcal{E}$ for all the tokens using the pre-trained features.\\
let $G'\leftarrow{G}$, $T'\leftarrow{\phi}$.\\
\While{Round in $[0,K)$}{
\textbf{Supervisor-step:}\\
\If{$T'\neq{\phi}$}{
Align the relations in $T'$ to $R$ with Eq. (\ref{eq:tras}).\\
$\Delta{T}\leftarrow$Find the top-$K$ triples in the aligned $T'$ with the trained $f_{G'}(*)$.\\
$T'\leftarrow\Delta{T}\bigcup{T'}$.}
Sample the negative triple set $N$ based on ${T'}$.\\
Train the KGE triple likelihood $f_{G'}(*)$ by minimizing Eq. (\ref{eq:kge_loss}). with $T'$ and $N$.\\
Sample the top-$k$ positive and negative entity pairs $P^+$ and $P^-$ based on Algorithm \ref{algorithm:sampling} with $T'$ and the embedding map $\mathcal{E}$.\\
\textbf{Explorer-step:}\\
Train the benchmark-based supervision JEE by minimizing the function in Eq. (\ref{eq:gjee}) with JEE training data.\\
Exhaustive generate the candidate triples $T'$ based on the mention results from the JEE testing data with the trained JEE.\\
}
\textbf{Output} $G'$.
}
\caption{Collaborative Knowledge Graph Fusion Algorithm}
\end{algorithm}
We initialize the embeddings for all tokens in the corpus with pre-trained features (BERT~\cite{DBLP:conf/naacl/DevlinCLT19} in this paper, but alternative methods could potentially be used if necessary). 
These embeddings are then used in the supervisor process to infer the positive or negative entity pair sets using a prior knowledge graph. Next, the obtained positive and negative entity pair sets are used to supervise the explorer process. Then the JEE model in the explorer process extracts improved entities and relations to enrich the prior knowledge graph. The supervisor adds the top-$K$ ranked aligned candidate triples in using beam search. 

\textbf{Discussion and Analysis.}
Our model links event extraction and knowledge graph fusion together as a single process.
This alternative process enhance the performance of both of the aforementioned tasks and also results a high quality enriched KG. The main reasons for these improvements are twofold. First, with more useful knowledge implications (evaluated extracted triples from the corpus) for a given knowledge graph, the semantic relationships between its entities are improved. As a result, the performance of the knowledge graph embedding with the enriched knowledge graph is also improved. Second, the accuracies for the entity and relation extraction tasks are also improved with 

 the help of the enriched knowledge graph.

\subsection{Negative Triple Sampling and Training}

Many existing methods use the randomized head or tail entity replaced triples from the positive triple set as the negative samples~\cite{DBLP:conf/iclr/KipfPW20}. To further improve the quality of the negative samples in Line 11 of Algorithm \ref{algorithm:em}, we treat the output of random sampled negative triples as the candidate set and then further use the KGE triple likelihood to measure their likelihoods. The final negative samples set in Line 11 of Algorithm \ref{algorithm:em} are the top-ranked samples from the candidate set based on the KGE triple likelihood scores.

\section{Experiments and Analysis}\label{sec:experiment}

In this section, we aim to address the following research questions: 
\begin{itemize}
  \item \textbf{RQ1:} Can a system in the proposed Collaborative Knowledge Graph Fusion framework successfully improve both the performances for the JEE and KGE tasks? 
  \item \textbf{RQ2:} Are the automatically extracted and translated triples valuable or suitable for the target KG?
  \item \textbf{RQ3:} What is the generalizability of a system with the proposed Collaborative Knowledge Graph Fusion framework representation across different real-world corpora and KGs? 
\end{itemize}
We also perform ablation analysis to investigate the effect of each module of the model, as well as a qualitative analysis of detailed examples.
\subsection{Datasets}
Since our system consists of the optimization processes of a JEE task and a KGF task, our dataset contains several real-world corpora for the JEE task and also two public Knowledge Graphs (KG) for the KGF task.

\textbf{The corpora.}
ACE 2005 \cite{ldc} is a widely used dataset to test the performances for the event extraction models. WebNLG is a corpus used for a challenge of natural language generation~\cite{DBLP:conf/acl/GardentSNP17}. CoNLL is a Spanish news corpus from~\cite{conll}. We create the NYT and CoNLL datasets\footnote{https://github.com/hkharryking/labeled\_NYT\_CoNLL} by preprocessing the original NYT~\cite{nyt} and CoNLL~\cite{conll} corpora with the CoreNLP\footnote{https://stanfordnlp.github.io/CoreNLP/}. This preprocessing includes annotating the triggers and entities from the sentences. 

\textbf{The Knowledge Graphs.} In order to implement the benchmark-based supervision mechanism at the explorer process, we preprocess the WN18 and FB15k-237 \cite{DBLP:conf/nips/BordesUGWY13} as the prior KGs in our tasks. Since the entities in each KG are encoded as the inner IDs, we map these IDs to the real entity mentions by the corresponding mapping files. Further, since freebase API depressed, we map the entity IDs in FB15k-237 to the URLs on the Wikidata\footnote{https://www.wikidata.org}) and then crawl the Wikidata titles to create the real entity mentions.

\textbf{Preprocessing Details.}
To implement a complete ``Collaborative Knowledge Graph Fusion'' framework, we preprocess the datasets to obtain the training sets and the testing sets for the supervisor and explorer respectively. The details of these preprocessed datasets are list in the following Tables.

\begin{table}[h]
	\centering
	\caption{Summary of the Corpora for the Explorer (JEE) Process}
	\vspace{-0.1in}
\begin{tabular}{l|cccc}
	\toprule
		 &ACE2005&NYT&CoNLL&WebNLG\\
	\midrule
Sentences&17,606&6,355&3,903&3,973\\
Training sent.&16,765&5,500&3,000&2,649\\
Testing sent.&841&855&903&1,324\\
	\bottomrule
	\end{tabular}
	\label{table:data:statistic:corpora}
\vspace{-0.1in}
\end{table}

\begin{table}[h]
	\centering
	\caption{Summary of the KGs for the Supervisor (KGF) Process}
\begin{tabular}{lccccc}
	\toprule
		 &&ACE2005&CoNLL&NYT&WebNLG\\
	\midrule
\multirow{2}{*}{FB15K}&Seed triples&20,00&3,440&3,000&3,973\\
&Testing triples&969&698&1,129&1,786\\
\multirow{2}{*}{WN18}&Seed triples&526&68&2,042&311\\
&Testing triples&129&68&730&113\\
	\bottomrule
	\end{tabular}
	\label{table:data:statistic:KGs}
\vspace{-0.1in}
\end{table}
\subsection{Comparison Baselines}

We provide the baselines on both the JEE and the KGF tasks. The performance of KGF is evaluated by the link prediction performance of the trained knowledge graph embedding.

\begin{table*}[hbt]
	\centering
	\caption{Detailed comparison on ACE 2005 testing set.}
	\label{table:ace:comparison}
	\scalebox{0.95}[0.95]{
    \begin{tabular}{l|cccccccccccc}
    	\toprule
      \multirow{2}*{Model}&\multicolumn{3}{c}{Event Trigger Identification}&\multicolumn{3}{c}{Event Trigger Classification}&\multicolumn{3}{c}{Event Argument Identification}&\multicolumn{3}{c}{Event Argument Classification}\\
      \cline{2-13}~&Precision&Recall&F1&Precision&Recall&F1&Precision&Recall&F1&Precision&Recall&F1\\
      \midrule
StagedMaxEnt&73.9&66.5&70.0&70.4&63.3&66.7&75.7&20.2&31.9&71.2&19.0&30.0\\
TwoStageBeam&76.6&58.7&66.5&74.0&56.7&64.2&74.6&25.5&38.0&68.8&23.5&35.0\\
Reranking&77.6&65.4&71.0&\textbf{75.1}&63.3&68.7&73.7&38.5&50.6&70.6&36.9&48.4
\\
Joint3EE&70.5&74.5&72.5&68.0&71.8&69.8&59.9&59.8&59.9&52.1&52.1&52.1\\
Seq2Seq&66.7&62.4&64.5&57.3&53.7&55.5&62.8&72.8&67.5&46.3&56.6&50.9\\
Seq2Seq$^*$&72.4&67.5&69.9&69.7&65.0&67.2&72.7&75.0&73.8&58.7&67.0&62.6\\
CRF$^*$&71.9&73.6&72.7&68.2&68.2&68.2&70.7&\textbf{79.6}&74.9&58.7&66.0&62.1\\
BERT&75.0&\textbf{75.0}&75.0&75.0&\textbf{75.0}&75.0&82.8&72.6&77.4& 71.4&69.0&70.2\\
\midrule
BJEE$_{wn18}$&88.9&66.7&76.2&85.7&60.0&70.6&88.2&77.8&82.7&80.4&72.6&76.3\\
BJEE$_{fb15k}$&\textbf{88.9}&72.7&\textbf{80.0}&\textbf{88.9}&72.7&\textbf{80.0}&\textbf{89.0}&77.7&\textbf{83.0}&\textbf{86.5}&\textbf{69.8}&\textbf{77.2}\\
    	\bottomrule
    \end{tabular}
    }
\vspace{-0.1in}
\end{table*}

\begin{table}[hbt]
	\centering
	\caption{Comparison on the entity extraction on the ACE2005 testing set.}
	\label{table:ace:eecomparison}
    \begin{tabular}{l|cccccccccccc}
    	\toprule
      {Model}&Precision&Recall&F1\\
      \midrule
Seq2Seq&67.5&83.2&74.6\\
Seq2Seq$^*$&74.4&85.1&79.4\\
CRF$^*$&75.2&84.6&79.6\\
Reranking&82.4&79.2&80.7\\
PipelineGRU&80.6&80.3&80.4\\
Joint3EE&82.0&80.4&81.2\\
BERT&89.2&78.3&83.4\\
\midrule
BJEE$_{wn18}$& 92.4&81.5&86.6\\
BJEE$_{fb15k}$&\textbf{95.1}&\textbf{83.0}&\textbf{88.6}\\
    	\bottomrule
    \end{tabular}
\vspace{-0.1in}
\end{table}
\textbf{JEE baselines.}

\begin{itemize}
\item StagedMaxEnt~\cite{DBLP:conf/naacl/YangM16} and TwoStageBeam~\cite{DBLP:conf/acl/LiJH13} are classic pipe-lined framework methods to extract the event factors jointly.

\item Reranking~\cite{DBLP:conf/naacl/YangM16} is the statistical state-of-the-art joint event extraction method.

\item Seq2Seq~\cite{DBLP:conf/aclnut/LimsopathamC16} is a Joint Event Extraction (JEE) model with the Sequence-to-Sequence framework. Our experiments use the universal Sequence-to-Sequence framework implementation from~\cite{wang2020crosssupervised}.
\item Seq2Seq$^*$~\cite{DBLP:conf/aclnut/LimsopathamC16} is the extended Seq2Seq model with the Glove~\cite{DBLP:conf/emnlp/PenningtonSM14} pre-trained features.
\item CRF$^*$~\cite{DBLP:conf/aclnut/LimsopathamC16} is a method extended from Seq2Seq with the conditional random field layer with the Glove~\cite{DBLP:conf/emnlp/PenningtonSM14} pre-trained features.
\item BERT~\cite{DBLP:conf/naacl/DevlinCLT19} is the original BERT with Seq2Seq downstream layers.
\item Joint3EE~\cite{DBLP:conf/aaai/NguyenN19} is an embedding-based method to extract the entities, event triggers and arguments together.
\item Benchmark-based Supervision JEE (BJEE) is the joint model proposed in our paper. Our model supervised by the benchmark entity pairs sampled from a given knowledge graph. It is the explorer process in Sec~\ref{sec:explorer}. The subscripts in the experimental results are the names of the given knowledge graphs.
\end{itemize}

\textbf{KGF baselines.}

\begin{itemize}
\item TransE~\cite{DBLP:conf/nips/BordesUGWY13} is a classic statistical KGF model. It assumes that a relation of a triple can be represented as the difference between the head and tail entity vectors of that triples. It trains the latent vectors for all the triples based on the aforementioned assumption.

\item ConvE~\cite{dettmers2018conve} is a KGF method to concatenate the vectors for entities to create a matrix to represent the triples. It applies the convolutional neural network to capture the proximity between entities in a triple.

\item Supervisor is the method proposed in our paper. It is the supervisor process in Sec~\ref{sec:supervisor} that iteratively enriches its training knowledge triples with the extracted result from the explorer process.
\end{itemize}

\subsection{Evaluation Metrics}

To compare the JEE and KGF tasks, we provide two families of metrics for them respectively.

\textbf{Supervisor process (KGF task).} In the KGF task, we apply the MRR, Hit@10, Hit@20 and Hit@30 as the metrics to measure the performance of a model to predict or judge the possibility of a triple. 

The MRR (Mean Reciprocal Rank, MRR) is computed by Equation in our work.
\begin{equation}
   MRR=\sum_{t\in{\hat{T}}}\frac{1}{rank_t},
\end{equation}
where $\hat{T}$ is the testing triple set for the testing process.

Hit@n is the ratio of the positive triples that contains in the top-$n$ ranked triples ($n=10,20,30$ in our experiment) by our models towards the testing triple set $\hat{T}$.

Since our system requires to run on the JEE and KGF tasks alternatively, in order to improve the efficiency we pre-sampled the positive and negative triples from the testing triples and wrote them to files. Our evaluation on the performances of the KGF tasks are based on these pre-sampled triples.

\textbf{Explorer process (JEE task).} The performance of JEE is measured by the Precision, Recall, and the F1-scores for the triggers, the entities, and the arguments. The Precision is measured by the ratio of the correct tags output by a model from all the tokens in a corpus and the Recall is the ratio of the predefined tags contains in the output tags of a model.

\subsection{Prototype System and Implementation Details}
We implement a prototype system with the proposed Collaborative Knowledge Graph Fusion framework with Pytorch. This system consists of an explorer process that performs the Joint-Event-Extraction (JEE) task to extract the triples from a corpus and a supervisor process that conduct the Knowledge Graph Fusion (KGF) process to train the KGE triple likelihood based on the prior Knowledge Graph (KG). As is introduced in Section~\ref{sec:framework}, our system enriches a prior KG as follows. In the beginning, the explorer process extracts the triples from a given corpus under the guidance (the Benchmark-based Supervision Mechanism) of the supervisor. After the explorer submits the triples to the supervisor, the supervisor translates the triples to suit the form of its prior KG. With the translated triples the supervisor assesses the quality of the triples based on the KGE triple likelihood (represents its own understanding of the prior KG). In the end, the supervisor merges high-quality triples found in the last step to its prior KG and it also updates the benchmarks to the explorer.

In order to create a fair comparison platform, all the sequence-to-sequence encoders were implemented based on a BERT~\cite{DBLP:conf/naacl/DevlinCLT19} of 768 hidden dimensions. Since our framework requires two alternative processes, we use an Adam optimizer~\cite{DBLP:journals/corr/KingmaB14} with $1e\text{-}3$ learning rate and 30 epochs to train the explorer process for non-BERT models and all the BERT-based models (include our own) are trained with $2e\text{-}5$ learning rate and 30 epochs. We apply an Adadelta~\cite{DBLP:journals/corr/abs-1212-5701} optimizer with $1e\text{-}1$ learning rate and 20 epochs to train the supervisor process. The rounds of the Collaborative Knowledge Graph Fusion framework are set to 8 for all our models. Both the weights for the benchmark-based supervision and the mention similarity ( $\alpha$ and $\gamma$) set to 0.5 in the prototype system. Besides, this prototype system runs on a Linux machine with 4 NVIDIA 2080TI GPUs.

\subsection{Comparison on the JEE task}

We compare our model with the others on the standard event extraction dataset ACE 2005. The results of the event trigger and argument extractions are shown in the Table~\ref{table:ace:comparison}. We observe that, the performances on all related subtasks of our model are superior to the other alternatives. We further compare the performance of the entity mention detection of our model with other methods, where our result also excels the other methods (in Table~\ref{table:ace:eecomparison}). All these results verify that effectiveness of the proposed supervisor-explorer mechanism 
boosts the performance of the JEE process. Besides, we find that, due to the sequence-to-sequence (seq2seq) uniform framework, the performances on the argument identification and classification tasks of the seq2seq-framework models are significant improved.

\begin{table*}[t]\small
	\centering
		\caption{Comparison on all the real-world datasets with overall performances.}
		\label{table:alldata:comparison}
    \begin{tabular}{ccccccccccccc}
    	\toprule
      \multirow{2}*{Model}&\multicolumn{3}{c}{ACE 2005}&\multicolumn{3}{c}{NYT}&\multicolumn{3}{c}{CoNLL}&\multicolumn{3}{c}{WebNLG}\\
      \cline{2-13}~&Precision&Recall&F1&Precision&Recall&F1&Precision&Recall&F1&Precision&Recall&F1\\
      \midrule
Seq2Seq$^*$&71.2&73.9&72.5&91.0&88.2&89.5&86.6&88.7&87.6&91.2&90.9&91.1\\
CRF$^*$&71.3&76.5&73.8&89.9&89.8&89.9&87.3&88.6&88.0&92.2&89.6&90.9\\
BERT&87.1&86.9&87.0&97.8&97.8&97.8&94.2&94.2&94.2&90.1&90.0&90.1\\
BJEE$_{fb15k}$&92.1&92.1&92.1&\textbf{99.2}&\textbf{99.2}&\textbf{99.2}&\textbf{96.3}&\textbf{96.3}&\textbf{96.3}&96.3&96.3&96.3\\
BJEE$_{wn18}$&\textbf{96.0}&\textbf{94.9}&\textbf{95.5}&99.0&99.0&99.0&95.8&95.6&95.7&\textbf{98.2}&\textbf{98.2}&\textbf{98.2}\\
    	\bottomrule
    \end{tabular}
\end{table*}

\begin{table*}[h]\small
	\centering
	\caption{Top extracted and aligned results from ACE 2005 corpus to knowledge graph FB15k by our system.}
	\label{table:case:study1}
  \resizebox{1\linewidth}{!}{	
  \begin{tabular}{cccccc}
  	\toprule
  
  		 \multirow{2}*{Rank}&\multicolumn{4}{c}{ACE 2005 corpus}&FB15K\\
  		 \cmidrule(r){2-5}\cmidrule(r){6-6}
  		 &Head Entity&Trigger mention&Trigger type&Tail Entity&Relation\\
  	\midrule
  	1&the Persian Gulf&killed&Life&all six British crew members&/people/deceased\_person/place\_of\_death\\
  	2&two Royal Navy helicopters&killed&Life&all six British crew members and one American&/people/cause\_of\_death/people\\
  	3&the capital&rained down&Conflict&aerial more than 300 Tomahawk cruise missiles&/people/deceased\_person/place\_of\_death\\
  	4&the United States&summit&Contact&the president Putin&/business/business\_operation/industry\\
  	5&the capital&took control&Baghdad the police stations&Movement&/location/country/form\_of\_government\\
  	\bottomrule
	\end{tabular}
	}
\end{table*}

To validate the universality of our method, we compare the overall extraction performances for the proposed JEE models guided by FB15K and WN18 knowledge graphs on all mentioned real-world datasets in the Table~\ref{table:alldata:comparison}. Since many methods do not consider these datasets, we only report the results of our implemented methods in this experiment. We can observe that the proposed method extracts better mentiones (both the event argument and trigger mentions) than the other non-knowledge-base-guided methods. Further, an interesting thing is that, although the CONLL is a Spanish corpus, the performances of the event extraction tasks on it can still be boosted by the proposed framework with the English-written knowledge graphs (FB15K and WN18). The reason is that many proper nouns are shared by both Spanish and English, and the semantic structure of them might also help the event extraction in Spanish. All the results in this experiment verify that the proposed Collaborative Knowledge Graph Fusion framework effectively boosts the performance of the JEE processes.

\subsection{Comparison on the KGF task}

We compare the performance of our method with the other KGF models on the triple prediction task in this experiment. This experiment conducts in the following way. The classic models TransE and ConvE are directly trained on the training set of the knowledge graph FB15K. The supervisor of our model is trained with an enriched training set that is obtained through the proposed Supervisor-explorer Collaborative Learning process. All models are tested with the same testing set of FB15K. The results of the supervisor model are obtained by alternatively run the supervisor and explorer processes to 8 rounds. Furthermore, since to enumerate all the negative triples requires weeks from our hardware platform, we only used 200 sampled negative triples with their corresponding positive triples as the testing set to compute the metrics. The result of this experiment is shown in Table~\ref{table:fb:kgc}.
\begin{table}[h]
	\centering
	\caption{Comparison on the KGF task on the FB15K.}
	\label{table:fb:kgc}
    \begin{tabular}{ccccc}
    	\toprule
      {Model}&Hit@10&Hit@20&Hit@30&MRR\\
      \midrule
TransE&100.0&70.0&60.0&0.0219\\
ConvE&100.0&100.0&93.3&0.0281\\
Supervisor&\textbf{100.0}&\textbf{100.0}&\textbf{100.0}&\textbf{0.0294}\\
    	\bottomrule
    \end{tabular}

\end{table}From Table~\ref{table:fb:kgc}, we observe that with the enriched triples, the performance of our KGF model is improved. This verifies that the obtained triples from our Collaborative Knowledge Graph Fusion framework bring useful information to predict the potential knowledge triples in a knowledge graph and the quality of the seed knowledge graph is enhanced.

\subsection{Ablation Analysis}
\begin{figure}[htb]
	\centering
		\includegraphics[width=3.0in]{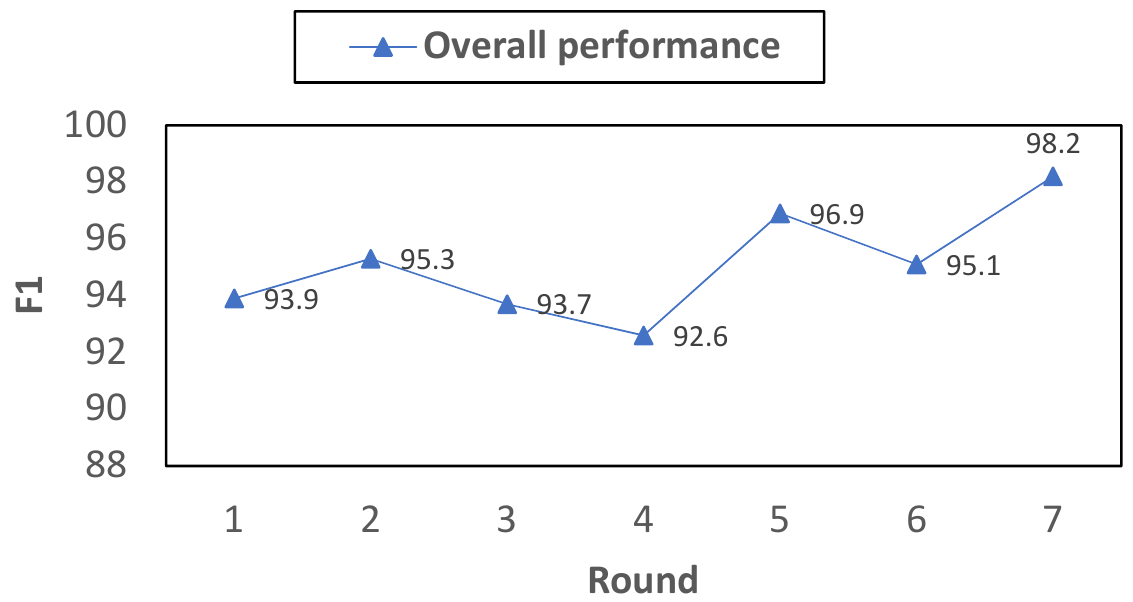}
	\caption{The overall extraction performance of the explorer process with different rounds (ran on the WebNLG corpus and supervised under the WN18 knowledge base).}
	\label{fig:epoch:ablation} 
  \vspace{-0.1in}
\end{figure}

Since we use BERT~\cite{DBLP:conf/naacl/DevlinCLT19} as the sequence-to-sequence encoder for our model, we compare the experimental results of our models (BJEE$_{wn18}$ and BJEE$_{fb15k}$) with the pure BERT~\cite{DBLP:conf/naacl/DevlinCLT19} model (with the same hidden dimensions) in Table~\ref{table:ace:comparison}, Table~\ref{table:ace:eecomparison} and Table~\ref{table:alldata:comparison}.
We observe that, with the proposed benchmark-based supervision mechanism, our results significantly outperform the pure BERT after the iterative learning process between the supervisor and explorer. To further discuss the influence of the iterative process, we also provide an experiment to compare the overall JEE performances with different iterative rounds in Figure~\ref{fig:epoch:ablation}.
From this figure, the overall JEE performance is improving with the iterative round increasing. This shows that the iterative process between the explorer and supervisor of our model indeed helps the overall performance of the JEE tasks.

\subsection{Sensitivity Analysis}
In order to further analyze the details of the proposed Collaborative Knowledge Graph Fusion framework, we provide several experiments to study the performance of our system with different forms of the teacher or explorer processes.

\begin{figure}[htb]
	\centering
		\includegraphics[width=3.0in]{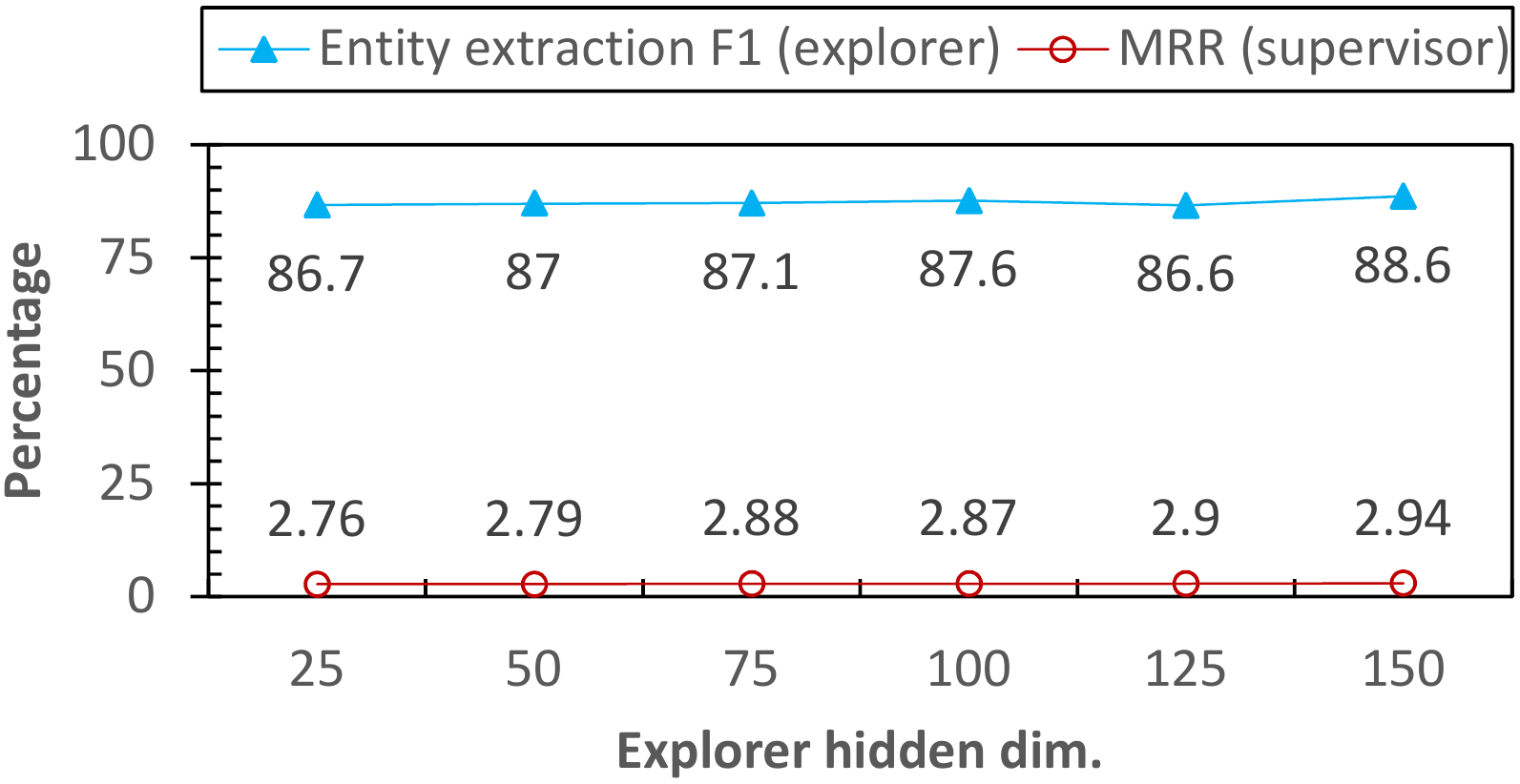}
	\caption{The performances of our system under different explorers.}
	\label{fig:ablation1} 
  \vspace{-0.1in}
\end{figure}

Figure~\ref{fig:ablation1} shows the performances of our system with a fixed teacher (with 4 CNN kernels) under explorers in differenct sizes of hidden dimensions. From
Figure~\ref{fig:ablation1}, we observe that with the same teacher, the diligent (with more hidden dimensions) of an explorer is, the better performance of the teacher process.
performs better.

\begin{figure}[htb]
	\centering
		\includegraphics[width=3.0in]{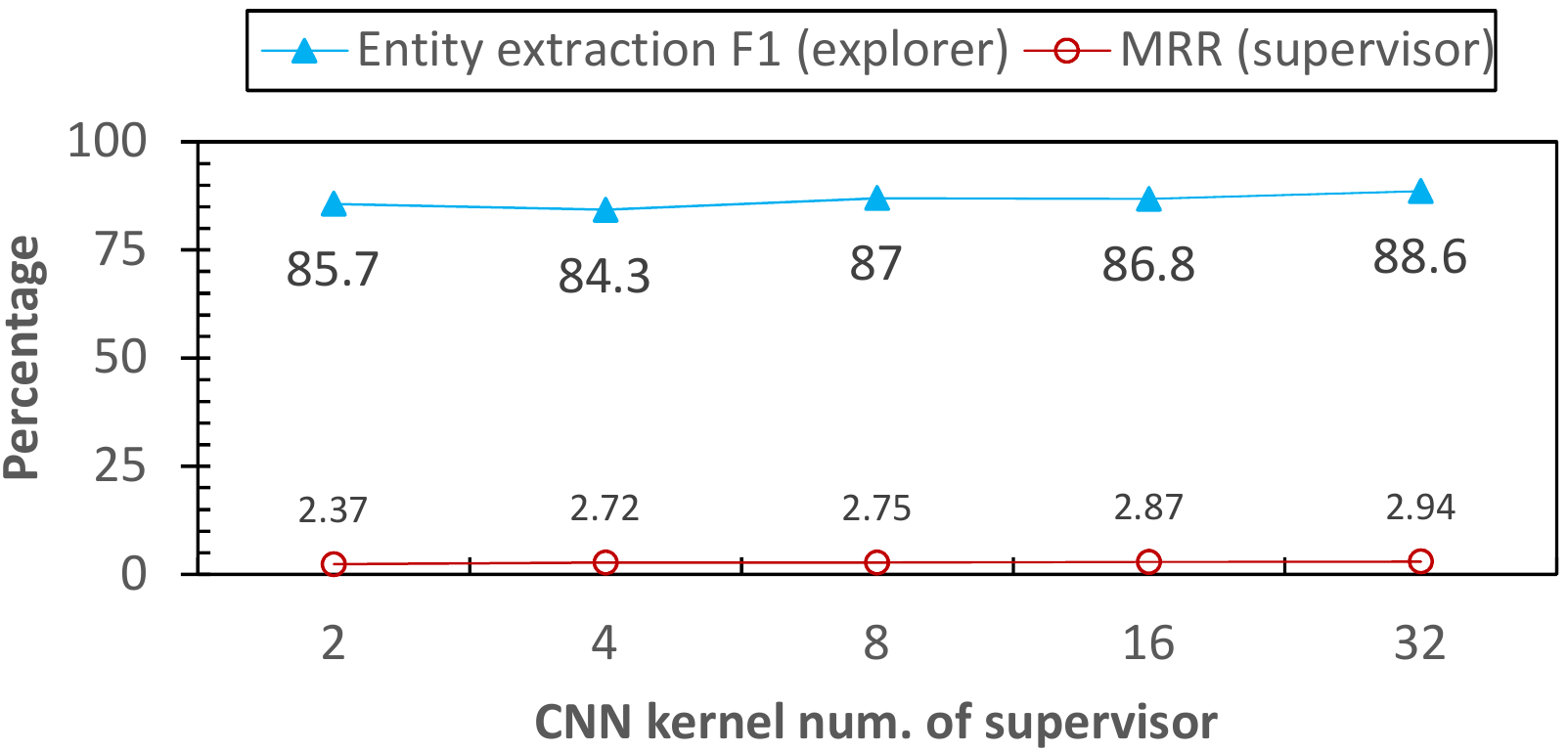}
	\caption{The performances of our system under different supervisors.}
	\label{fig:ablation2} 
  \vspace{-0.1in}
\end{figure}

Figure~\ref{fig:ablation2} gives the performances of our system with a fixed explorer (with 150 hidden dimensions) under supervisors in different numbers of CNN kernels. From this figure, we observe that, with the same explorer, the performance of our system peaks with the supervisor having a certain number (32 in this expeiment) of CNN kernels.

The two aforementioned experiments indicates that, the overall performance of a system with the SSL framework might be boosted by improving the explorer process, and the improvement of this overall performance is limit with the same explorer under different supervisors.

\subsection{Case study: Translate and Align the Triples}

As is introduced in Algorithm~\ref{algorithm:em}, the explorer process of our system extracts new triples from the given corpus (ACE 2005) and generates a mapper to align the relations of these extracted triples to the relations in the knowledge graph (FB15K). Then, with the aligned relation mapper, our prototype system translates all the extracted triples in the forms of the target knowledge graph. In the last step, the explorer process ranks these translated triples with the trained KGE likelihood function from the supervisor and submits the top triples to the supervisor.

To further analyze the detail performance of the proposed TRAS (Translated Relation Alignment Score) method, we explore the automatically aligned relations by our Collaborative Knowledge Graph Fusion framework in the task to explore (extract) the ACE 2005 corpus guided by the FB15K knowledge graph.
 
We pick some top-ranked aligned and translated triples from the ACE 2005 corpus by our system and list them in Table~\ref{table:case:study1}. We can observe that most of these triples are aligned to the suitable relations in FB15K based on the given corpus. For example, our system aligns and the trigger mention ``killed'' of the type ``Life'' to the FB15K relation ``/people/deceased\_person/place\_of\_death'' for the 1-st triple extracted from the ACE 2005 corpus. In this result, our system infers that the trigger mention ``killed'' of the ACE 2005 corpus is highly similar to the relation ``/people/deceased\_person/place\_of\_death'' of the knowledge graph FB15K. In this result, our system infers that the trigger mention ``killed'' of the ACE 2005 corpus is aligned to the relation ``/people/deceased\_person/place\_of\_death'' of the knowledge graph FB15K. Our system makes this inference by considering both the semantic similarity between the mentions `killed'' and ``deceased'' and the affinities of the ``PER'' entities around the corresponding relations in the two sources. This shows that the proposed TRAS score provides a possible way for the fully-automatically knowledge graph fusion of the future works.



\section{Related Works}\label{sec:related_works}
In this section, we survey the related works to ours from the perspectives of joint event extraction, knowledge graph fusion and open information extraction.

\subsection{Joint Event Extraction}

Joint event extraction (JEE) aims to obtain the named entities, trigger mentions and relations simultaneously from a given corpus. Many recent works apply the pipe-lined method to achieve this task. That is to train a series of classifiers for the aforementioned sub-tasks and classify the mentions in sentences as different triggers at first. Then, with the classified triggers to identify the entity mentions or relations. StagedMaxEnt~\cite{DBLP:conf/naacl/YangM16} and TwoStageBeam~\cite{DBLP:conf/acl/LiJH13} are such kind pipe-lined systems. Reranking~\cite{DBLP:conf/naacl/YangM16} is the state-of-the-art statistical pipe-lined method for the JEE task.

Most neural network models apply the embedding method to capture the latent semantic relationships between sentence tokens and try to train different classifiers for different sub-tasks. Joint3EE~\cite{DBLP:conf/aaai/NguyenN19} is a such model with the multitask learning framework. However, since the separate training for different classifiers increases the sparsity of the efficient samples to each single classifier, the performance improvement of these methods are limited. The sequence-to-sequence methods~\cite{wang2020crosssupervised} train a neural network model to match a sentence in forms of a token sequence to a tag sequence. This kind of method focuses all sub-tasks to a single classifier and thus further improves the performance with the limited training data.

\subsection{Knowledge Graph Fusion}

Knowledge graph fusion~\cite{DBLP:journals/inffus/NguyenVJ20} is a task to fuse a knowledge graph with other data sources. Many KGF systems apply an ``enumerate-and-rank'' framework~\cite{DBLP:conf/ijcai/WangWG15} to complete a knowledge graph. That is, to train a classifier based on a given knowledge graph and identify the possible triples from a series of candidate triples. Usually, such classifer is based on the knowledge graph embedding (KGE)~\cite{DBLP:journals/tkde/WangMWG17} method. The TransE~\cite{DBLP:conf/nips/BordesUGWY13} is a classic KGE method to learn the embedding vectors to represent the triples in a knowledge graph. Recently, many works apply the neural network method to improve the performance of the KGE task. ConvE~\cite{dettmers2018conve} is a neural network KGE model with the convolutional neural network modules. As far as we know, none of the existing methods considers to link the JEE task to the KGE to create an automatically Knowledge Graph Fusion with Open Corpus.

\subsection{Open Information Extraction (Open IE)}
Open Information Extraction (Open IE)~\cite{DBLP:conf/acl/AngeliPM15} is another way to generate structural information from text sources. The traditional methods~\cite{DBLP:journals/corr/StanovskyFDG16,DBLP:conf/ijcai/Mausam16} get the new relation facts to form a KG based on the hand-crafted patterns. Recent works \cite{DBLP:conf/acl/CuiWZ18,DBLP:conf/acl/TrisedyaWQZ19} apply the neural relation extraction methods to directly generate relational facts from a given corpus and integrate them to an existing KG. During the integration process, these works trained a classifier to judge the correctness of the obtained relations according to the given KG. However, although the current Open IE works extract relational facts (triples) directly from text sources, few of them discuss that how to automatically merge the obtained facts to create a uniform and high-quality KG.

\section{Conclusion and Future Work}\label{sec_conclusion}
This paper has proposed a novel Collaborative Knowledge Graph Fusion framework to integrate the joint event extraction and the knowledge graph fusion tasks together. The implemented prototype system with the proposed framework could both extract the entity and trigger mentions and enrich the extracted mentions to a knowledge graph in the form of the knowledge graph triple (entity-relation-entity). To this end, we propose the benchmark-based supervision mechanism to guide the event extraction process of our system with a given knowledge graph and our system also merges the extracted triples to the target knowledge graph by referring the proposed Translated Relation Alignment Score. We test our prototype system on several real-world corpora and knowledge graphs. The experimental results show that our method improves the performances of both the event extraction and knowledge graph fusion processes after the alternatively training. Moreover, the aligned and translated relations from our system also show good interpretability about the improvements of the performances.
Our future work is to align the triples directly with their semantic meanings to further improve the performance of our model.




%
\bibliographystyle{IEEEtran}
\bibliography{references}

%
\vspace{-0.71in}
\begin{IEEEbiography}[{\includegraphics[width=1in,height=1.25in,clip,keepaspectratio]{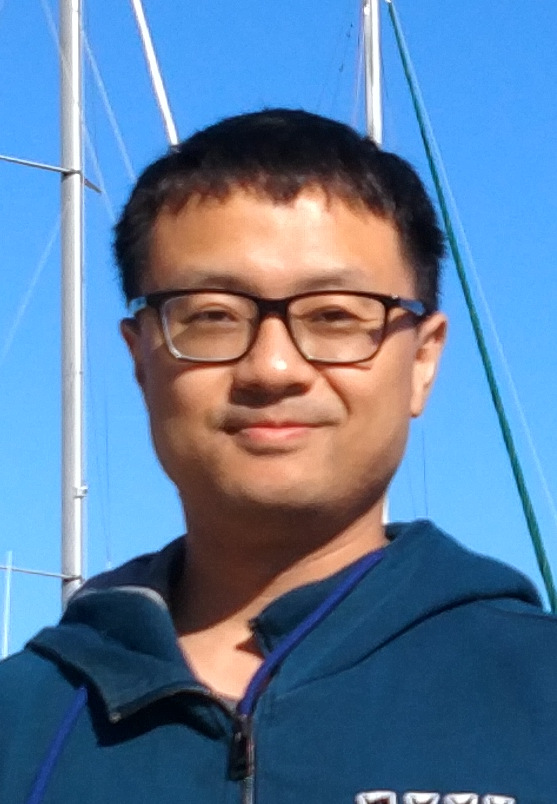}}]{Yue Wang} received the Ph.D. degree from Sichuan University, Sichuan, China. He was a postdoctor of Peking University, Beijing, China. He is now an Associate Professor at Central University of Finance and Economics, Beijing, China. He has published more than 30 journal and conference papers, including TKDE, WWWJ, Science China: Information Science, IJCAI, ICDM, IEEE BigData etc. His current research interests include data mining and machine learning.
\end{IEEEbiography}
\vspace{-0.71in}
\begin{IEEEbiography}[{\includegraphics[width=1in,height=1.25in,clip,keepaspectratio]{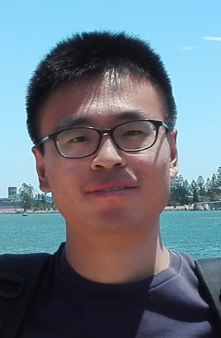}}]{Yao Wan}
received his Ph.D degree from the College of Computer Science, Zhejiang University, Hangzhou, China, in 2019. He is currently a lecturer of the College of Computer Science and Technology, Huazhong University of Science and Technology. 
He has been a visiting student of University of Technology Sydney and University of Illinois at Chicago in 2016 and 2018, respectively.
His research interests lie in the synergy between artificial intelligence and software engineering, especially natural language processing, programming languages, software engineering, and machine learning.
\end{IEEEbiography}
\vspace{-0.7in}
\begin{IEEEbiography}[{\includegraphics[width=1in,height=1.25in,clip,keepaspectratio]{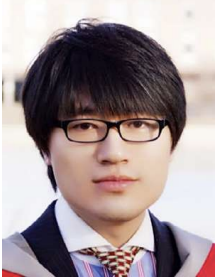}}]{Lu Bai} received the Ph.D. degree from the
University of York, UK, and both the B.Sc. and
M.Sc degrees from Macau University of Science
and Technology, Macau SAR, China. He was a
recipient of the National Award for Outstanding
Self-Financed Chinese Students Study Aboard by
China Scholarship Council in 2015, and the Best Paper Awards of the International Conferences ICIAP
2015 (Eduardo Caianello Best Student Paper Award)
and ICPR 2018. He is now a Professor
 in School of Artificial Intelligence, Beijing Normal University, Beijing, China,
Beijing, China. He has published more than 80 journal and conference papers, including TPAMI, TNNLS, TCYB, PR, ICML, IJCAI, ECML-PKDD, ICDM, etc. His current research interests include pattern recognition, machine
learning, and financial data analysis. He is currently a member of the editorial board of the journal Pattern Recognition
\end{IEEEbiography}
\vspace{-0.7in}
\begin{IEEEbiography}[{\includegraphics[width=1in,height=1.25in,clip,keepaspectratio]{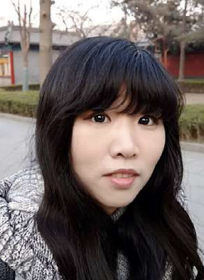}}]{Lixin Cui} received the Ph.D. degree from the University of Hong Kong, HKSAR, China, and both the B.Sc. and M.Sc. degrees from Tianjin University, Tianjin, China. She is now an Associate Professor at Central University of Finance and Economics, Beijing, China. She was the recipient of the Outstanding Paper Awards of the International Conference IEEE IEEM 2019, the Best Student Paper Awards of the International Conferences
APIEMS 2011 and WCE 2011. She is currently an Associate Editor of Pattern Recognition Journal. She has published more than 40 journal and conference papers, including
TPAMI, TFS, TCYB, TNNLS, PR, WWWJ, IJCAI, ECML-PKDD, etc. Her current research interests include machine learning, deep learning, and their applications in Fintech problems. She is currently a member of the editorial
board of the journal Pattern Recognition.
\end{IEEEbiography}
\vspace{-0.7in}
\begin{IEEEbiography}[{\includegraphics[width=1in,height=1.25in,clip,keepaspectratio]{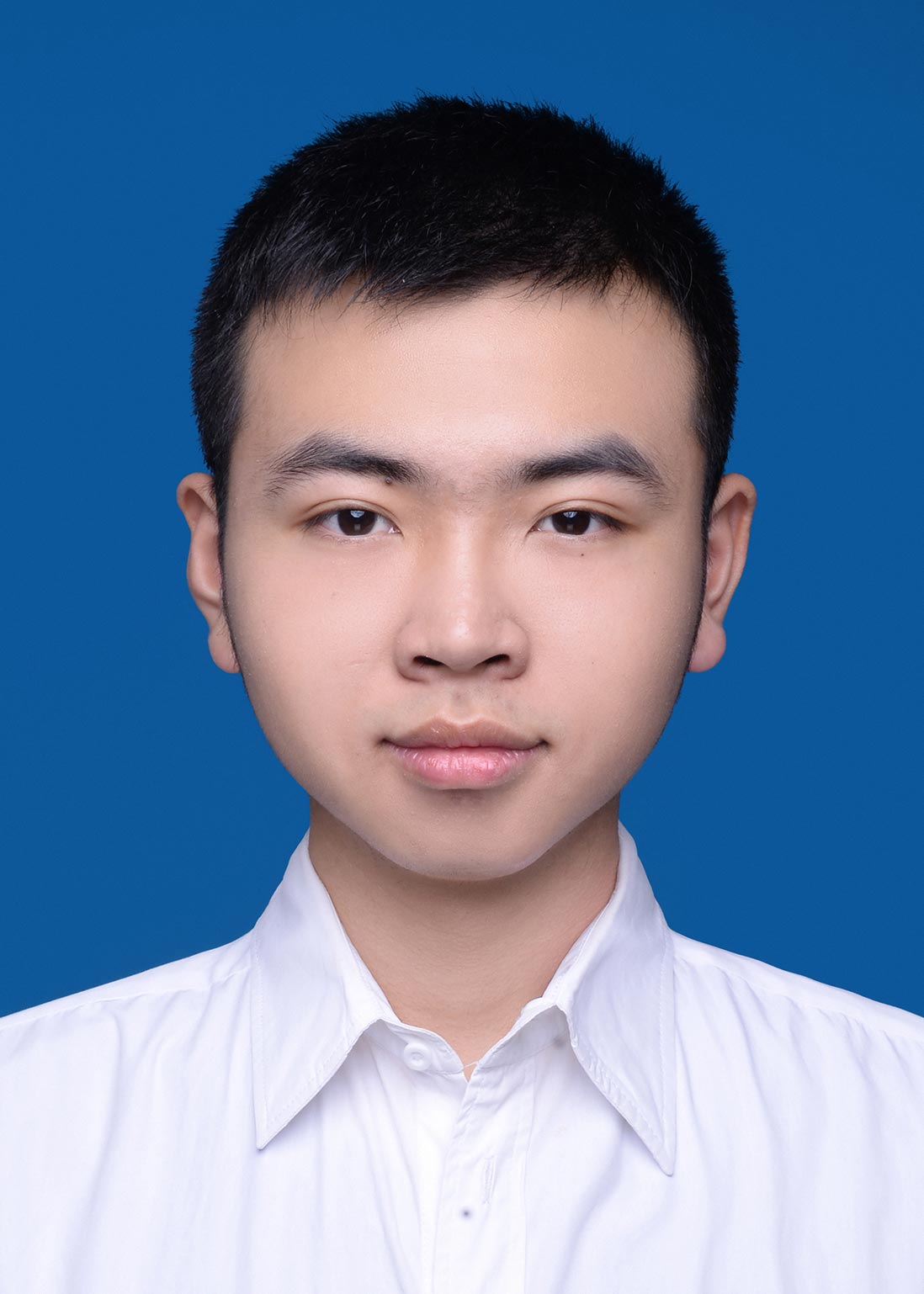}}]{Zhuo Xu} received the B.Sc. degrees from Central University of Finance and Economics. He is now an graduate explorer in Central University of Finance and Economics, Beijing, China.
\end{IEEEbiography}
\vspace{-0.7in}
\begin{IEEEbiography}[{\includegraphics[width=1in,height=1.25in,clip,keepaspectratio]{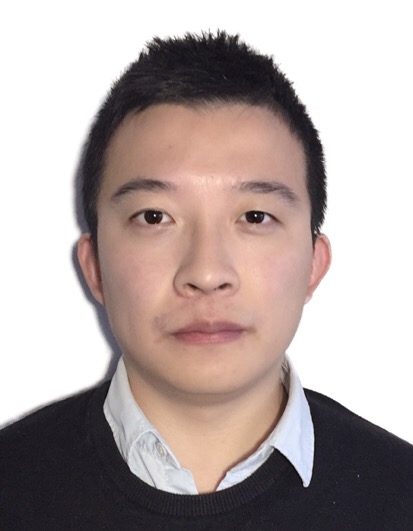}}]{Ming Li}is currently a ''Shuang Long Scholar'' Distinguished Professor at the Key Laboratory of Intelligent Education Technology and Application of Zhejiang Province, Zhejiang Normal University, China. He received his PhD degree from the Department of Computer Science and IT at La Trobe University, Australia. He completed two Postdoctoral Fellowship positions with the Department of Mathematics and Statistics, La Trobe University, Australia, and the Department of Information Technology in Education, South China Normal University, China, respectively. He has published in top-tier journals and conferences, including Artificial Intelligence, IEEE TCYB, IEEE TII, ACM TMOS, NeurIPS, ICML. He, as a leading guest editor, organized a special issue ``\emph{Deep Neural Networks for Graphs: Theory, Models, Algorithms and Applications}'' in IEEE TNNLS. He is a PC member at ICML, AAAI, NeurIPS, ICLR, AJCAI, KDD, and an Associated Editor of Neural Networks.
\end{IEEEbiography}
\vspace{-0.7in}
\begin{IEEEbiography}[{\includegraphics[width=1in,height=1.25in,clip,keepaspectratio]{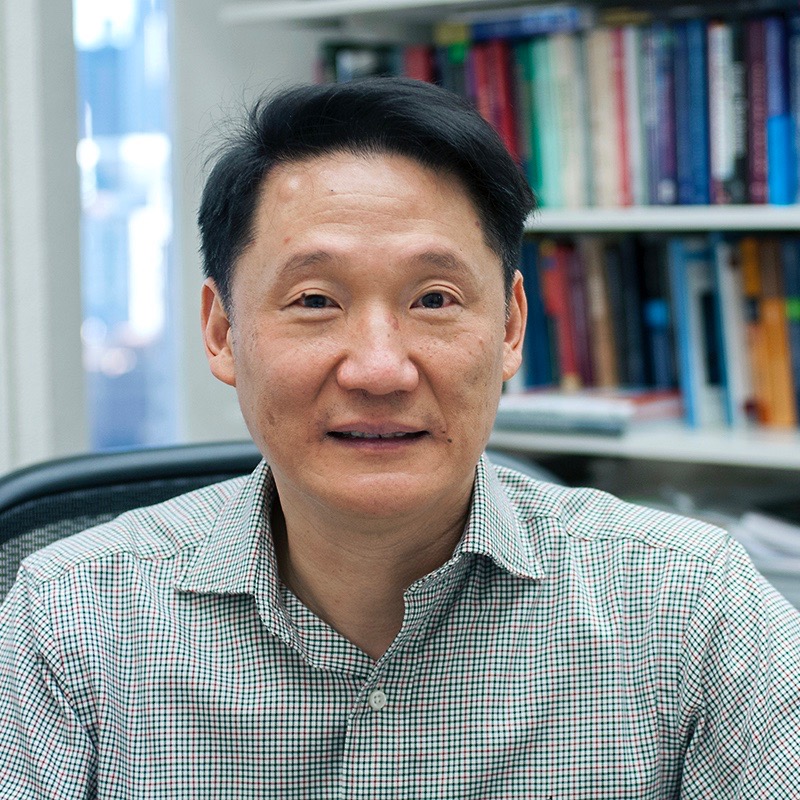}}]{Philip S. Yu} received received the B.S. degree in electrical engineering from National Taiwan University, the M.S. and Ph.D. degrees in EE from Stanford University, and the MBA degree from New York University. He is currently a Distinguished Professor of computer science with the University of Illinois at Chicago (UIC), and holds the Wexler Chair in information technology. He has published more than 970 papers in refereed journals
and conferences. He holds or has applied for over 300 US patents. He was a member of the Steering Committee of the IEEE Data Engineering and the IEEE Conference on Data Mining. He is a Fellow of the ACM and the IEEE. He is on the Steering Committee of the ACM Conference on Information and Knowledge Management. He received the ACM SIGKDD 2016 Innovation Award for his influential research and scientific contributions on mining, fusion, and anonymization of big data, the IEEE Computer Society’s 2013 Technical Achievement Award for ”pioneering and fundamentally innovative contributions to the scalable indexing, querying, searching, mining, and anonymization
of big data”, and the Research Contributions Award from ICDM 2003, for his pioneering contributions to the field of data mining. He also received the ICDM 2013 10-year Highest-Impact Paper Award, and the EDBT Test of Time Award (2014). He has received several IBM honors, including two IBM Outstanding Innovation Awards, an Outstanding Technical Achievement Award, two Research Division Awards, and the 94th plateau of Invention
Achievement Awards. He was the Editor-in-Chief of the IEEE Transactions on Knowledge and Data Engineering (2001-2004).
\end{IEEEbiography}
\vspace{-0.7in}
\begin{IEEEbiography}[{\includegraphics[width=1in,height=1.25in,clip,keepaspectratio]{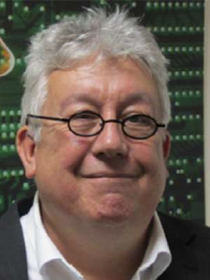}}]{Edwin R. Hancock} received the B.Sc., Ph.D., and D.Sc. degrees from the University of Durham, Durham, UK. He is currently an Emeritus Professor with the Department
of Computer Science, University of York, York, UK. He has published over 200 journal articles and 650 conference papers. Prof. Hancock was a recipient of the Royal Author Biography Society Wolfson Research Merit Award in 2009, the Pattern Recognition Society Medal in 1991, the BMVA Distinguished Fellowship in 2016 and the IAPR Piere Devijver Award in 2018. He is a fellow of the IAPR, IEEE, the Royal Astronomical Society, the Institute of Physics, the Institute of Engineering and Technology, and the British Computer Society. He was named Distinguished Fellow by the British Machine Vision Association. He has also received best paper prizes at CAIP 2001, ACCV 2002, ICPR in 2006 and 2018, BMVC 20 07, ICIAP in 2009 and 2015. He is currently Editor-in-Chief of the journal Pattern Recognition, and was founding Editor-in-Chief of IET Computer Vision from 2006 until 2012. He has also been a member of the editorial boards of the journals IEEE Transactions on Pattern Analysis and Machine Intelligence, Pattern Recognition, Computer Vision and Image Understanding, Image and Vision Computing, and the International Journal of Complex Networks. He has been Conference Chair for BMVC in 1994 and Program Chair in 2016, Track Chair for ICPR in 2004 and 2016 and Area Chair at ECCV 2006 and CVPR in 2008 and 2014, and in 1997 established the EMMCVPR workshop series. He was Second Vice President of the International Association of
Pattern Recognition (2016-2018). He is currently an IEEE Computer Society Distinguished Visitor (2021-2023).
\end{IEEEbiography}




\end{document}